\documentclass[conference]{IEEEtran}
\IEEEoverridecommandlockouts
\usepackage{cite}
\usepackage{amsmath,amssymb,amsfonts}
\usepackage{algorithmic}
\usepackage{graphicx}
\usepackage{textcomp}
\usepackage{xcolor}
\usepackage{lipsum}
\usepackage{amsmath}
\usepackage{amssymb}
\usepackage{graphicx}
\usepackage{algorithmic}
\usepackage{adjustbox}
\usepackage{bm}
\usepackage{resizegather}
\usepackage[ruled, linesnumbered, longend]{algorithm2e}
\def\BibTeX{{\rm B\kern-.05em{\sc i\kern-.025em b}\kern-.08em
    T\kern-.1667em\lower.7ex\hbox{E}\kern-.125emX}}
\begin{document}

\title{\textbf{Fine-tuning BERT with Bidirectional LSTM for Fine-grained Movie Reviews Sentiment Analysis  \thanks{Extend version of  Sentiment Analysis of Movie Reviews Using BERT, presented at The Fifteenth International Conference on Information, Process, and Knowledge Management, eKNOW23, Venice, Italy, 2023. } }}
\author{\IEEEauthorblockN{Gibson Nkhata, Susan Gauch}
\IEEEauthorblockA{\textit{Department of Computer Science \& Computer Engineering} \\
\textit{University of Arkansas}\\
Fayetteville, AR 72701, USA \\
Email: gnkhata@uark.edu, sgauch@uark.edu}
\and
\IEEEauthorblockN{Usman  Anjum, Justin Zhan}
\IEEEauthorblockA{\textit{Department of Computer Science} \\
\textit{University of Cincinnati}\\
Cincinnati, OH 45221, USA \\
Email: anjumun@ucmail.uc.edu, zhanjt@ucmail.uc.edu}

}

\maketitle

\begin{abstract}
  Sentiment Analysis (SA) is instrumental in understanding people’s viewpoints, facilitating social media monitoring, recognizing products and brands, and gauging customer satisfaction. Consequently, SA has evolved into an active research domain within Natural Language Processing (NLP).  Many approaches outlined in the literature devise intricate frameworks aimed at achieving high accuracy, focusing exclusively on either binary sentiment classification or fine-grained sentiment classification. In this paper, our objective is to fine-tune the pre-trained BERT model with Bidirectional LSTM (BiLSTM) to enhance both binary and fine-grained SA specifically for movie reviews. Our approach involves conducting sentiment classification for each review, followed by computing the overall sentiment polarity across all reviews. We present our findings on binary classification as well as fine-grained classification utilizing benchmark datasets. Additionally, we implement and assess two accuracy improvement techniques, Synthetic Minority Oversampling Technique (SMOTE) and NLP Augmenter (NLPAUG), to bolster the model's generalization in fine-grained sentiment classification. Finally, a heuristic algorithm is employed to calculate the overall polarity of predicted reviews from the BERT+BiLSTM output vector. Our approach performs comparably with state-of-the-art (SOTA) techniques in both classifications. For instance, in binary classification, we achieve 97.67\% accuracy, surpassing the leading SOTA model, NB-weighted-BON+dv-cosine, by 0.27\% on the renowned IMDb dataset. Conversely, for five-class classification on SST-5, while the top SOTA model, RoBERTa+large+Self-explaining, attains 55.5\% accuracy, our model achieves 59.48\% accuracy, surpassing the BERT-large baseline by 3.6\%.  
\end{abstract}


\section{Introduction}

This paper builds upon our prior research~\cite{gnkhata}, into sentiment analysis (SA) of movie reviews using Bidirectional Encoder Representations from Transformers (BERT)~\cite{devlin2018bert} and computing overall polarity within the scope of binary sentiment classification. Still, SA aims to discern the polarity of emotions (e.g., happiness, sorrow, grief, hatred, anger, and affection) and opinions derived from text, reviews, and posts across various media platforms~\cite{baid2017sentiment}. It plays a crucial role in gauging public sentiment, serving as a potent marketing tool for comprehending customer emotions across diverse marketing campaigns. SA significantly contributes to social media monitoring, brand recognition, customer satisfaction, loyalty, advertising effectiveness, and product acceptance. Consequently, SA stands as one of the most sought-after and impactful tasks within Natural Language Processing (NLP)~\cite{mesnil2014ensemble}. It encompasses polarity classification, involving binary categorization, and fine-grained classification, encompassing multi-scale sentiment distribution.

Movie reviews serve as a crucial means to evaluate the performance of a film. While assigning a numerical or star rating offers a quantitative assessment of a movie's success or failure, a collection of movie reviews offers a qualitative exploration of various aspects within the film. Textual movie reviews provide insights into strengths and weaknesses of a movie, allowing for a deeper analysis that gauges the overall satisfaction of the reviewer. This study focuses on SA of movie reviews due to the availability of standardized benchmark datasets and the existence of significant qualitative works within this domain, as highlighted in publications such as~\cite{bingyu2022document}.

 BERT stands as a renowned pre-trained language representation model, demonstrating strong performance across various NLP tasks such as named entity recognition, question answering, and text classification~\cite{devlin2018bert}. Its versatility extends to information retrieval, where it has been leveraged to construct efficient ranking models for industry-specific applications~\cite{guo2020detext}. Additionally, its adaptability is evident in applications like extractive summarization of text, as successfully demonstrated in~\cite{liu2019fine}, and in question answering tasks, where it yielded satisfactory results as seen in~\cite{he2020infusing}. The model's efficacy was further highlighted in data augmentation strategies, leading to superior outcomes, as exemplified in~\cite{yang2019data}. While BERT has found primary use in SA~\cite{munikar2019fine}, its accuracy on certain datasets remains a challenge. 

 Ambiguity in NLP, particularly SA, arises due to the complexity of language. Words and phrases often carry multiple meanings or interpretations based on context, making it challenging to accurately discern sentiment. This challenge is evident in instances where words might have different connotations depending on their context within a sentence or across diverse texts. BERT addresses this issue by leveraging its bidirectional context understanding~\cite{devlin2018bert}. Unlike earlier models that processed language in one direction, BERT comprehends words based on both preceding and succeeding words, allowing it to capture a more nuanced understanding of context. Therefore, BERT can better grasp the intricate meanings and resolve ambiguities present in natural language, thereby improving SA accuracy.
 
Bidirectional Long Short-Term Memory (BiLSTM) networks are also very popular for text classification in NLP ~\cite{siami2019performance}. BiLSTM is beneficial in SA due to its ability to capture contextual information from both past and future inputs in long sequences~\cite{graves2005framewise}. In SA, context is crucial, words derive their meaning not just from preceding words but also from the subsequent ones. BiLSTMs excel in capturing this bidirectional context by processing sequences in two directions simultaneously: forward (from the beginning to the end of a sequence) and backward (from the end to the beginning). This capability enables BiLSTMs to model more nuanced and complex dependencies in text, leading to improved SA performance compared to unidirectional models.

Fine-tuning is a common technique for transfer learning. The target model copies all model designs with their parameters from the source model except the output layer and fine-tunes these parameters based on the target dataset~\cite{devlin2018bert}. The main benefit of fine-tuning is no need to train the entire model from scratch, reducing the training time of the target model. Hence, we are fine-tuning BERT by coupling BiLSTM and training the model on movie reviews SA benchmarks. BERT, with its attention mechanisms and bidirectional context understanding, captures rich contextual information~\cite{devlin2018bert}. Combining it with a BiLSTM enhances this capability further. The ability of the BiLSTM to retain long-range dependencies complements contextual understanding of BERT, leading to a more nuanced comprehension of sentiment in text. The efficacy of this amalgamation is corroborated by conducting an ablation study in our experiments.

In our approach, we derive an overall polarity from the output vector generated by BERT+BiLSTM, employing a heuristic algorithm adapted from~\cite{arasteh2021will}. This algorithm is tailored uniquely in our study, addressing the specifics of the output vectors derived from binary, three-class, four-class, or five-class sentiment classifications. As a result, our work implements four distinct iterations of the algorithm, each corresponding to one of the four different sentiment classification tasks undertaken in this research.

Previous studies have predominantly focused on either binary sentiment classification or fine-grained SA, rarely combining both aspects. This paper addresses this gap by presenting an approach that fine-tunes BERT specifically for SA on movie reviews. Our objective is to conduct a comparative analysis encompassing both binary and fine-grained sentiment classifications. Through the integration of BERT and BiLSTM architecture, our fine-tuning methodology caters to both binary and fine-grained sentiment classification tasks. Notably, our approach surpasses state-of-the-art (SOTA) models in accuracy using our best-performing method. To address the challenge of class imbalance in fine-grained classification, we implement oversampling and data augmentation techniques on the respective datasets before feeding the data into the model classifier.

The main contributions in this work are as follows:
\begin{itemize}
\item 
Fine-tune BERT by coupling it with BiLSTM for both binary and fine-grained sentiment classification on well-known benchmark datasets and achieve accuracy that surpasses SOTA models.
\item 
Refine techniques to enhance the model's accuracy in fine-grained sentiment classification.
\item
Compute the overall sentiment polarity of predicted reviews based on the output vector from BERT+BiLSTM.
\item
Compare and evaluate our experimental outcomes against the results obtained from other studies, including those from SOTA models, using benchmark datasets. 
\end{itemize}

This paper is organized as follows: Section \ref{RelatedWork} describes the related work, Section \ref{Methodology} details the methodology, Section \ref{Experiments} discusses the experiments and results, and finally, Section \ref{concl} presents the conclusion and discusses future work.

\section{Related work} \label{RelatedWork}
This section covers relevant literature concerning SA, the intersection of deep learning and SA, particularly focusing on movie reviews SA, and the role of BERT in SA.

\subsection{SA}	
SA within NLP remains an active area of research.~\cite{anandarajan2019sentiment} introduced a step-by-step lexicon-based SA method using the R open-source software. The study conducted polarity classification on 1,000 movie reviews from the IMDb dataset, achieving an accuracy of 81.30\% by evaluating built-in lexicons.

In a contrasting approach,~\cite{baid2017sentiment} explored traditional machine learning techniques, i.e., Naive Bayes (NB), K-Nearest Neighbours (KNN), and Random Forests, for SA on IMDb movie reviews. Their findings highlighted NB as the top performer, achieving an accuracy of 81.45\%.

A different route was taken in~\cite{mesnil2014ensemble}, deploying an ensemble generative technique across multiple machine learning approaches for movie reviews SA, achieving an accuracy of 90.57\%. Conversely,~\cite{daeli2020sentiment} focused on the Cornell movie review dataset, solely utilizing KNN with the information gain technique, yielding an accuracy of 90.8\%. These studies underscore the effectiveness of KNN in traditional machine-learning methods for movie reviews SA.

A novel approach by training document embeddings using cosine similarity and feature combination with NB weighted bag of \textit{n}-grams was proposed in~\cite{thongtan2019sentiment}. Their comparison between training document embeddings using cosine similarity and dot product favored cosine similarity, achieving an accuracy of 91.42\% on the IMDb dataset.
In addition,~\cite{singh2020revisiting} applied mixed objective function for binary classification on SA on IMDb benchmark. Their approach reported an error rate of 9.95\%. The aforementioned models targeted polarity or binary sentiment classification only. Nevertheless, both binary classification and fine-grained classifications on SA were implemented in the following two studies.~\cite{semwal2018practitioners}  used transfer learning, while~\cite{wang2021entailment}, utilised entailment and few-shot learning. Both studies used IMDb, SST-2, and MR benchmarks for binary classification, and Yelp and SST-5 for fine-grained classification. Average accuracy of 87.57\%  is reported in~\cite{semwal2018practitioners} on binary datasets, while~\cite{wang2021entailment} has reported 88.16\%. For fine-grained classification, they reported average accuracy of 52.65\% and 54.65\%, respectively.

In our work, we adopt both polarity and fine-grained classifications from~\cite{wang2021entailment} but use deep learning techniques and the BERT pre-trained language model. We also adopt transfer learning from~\cite{semwal2018practitioners}.

\subsection{Deep learning}
This section explores the realm of deep learning applied to both the general SA task and specifically to movie reviews SA.

\subsubsection{Deep Learning on SA}
Deep learning stands as a SOTA technique for many NLP tasks, including SA. In a study by~\cite{zhang2015character}, Character-level Convolutional Neural Networks (CCNNs) were explored for text classification using Yelp and Amazon benchmarks. These CCNNs were compared against bag-of-words, n-grams, their TF-IDF variants, word-based CNNs, and Recurrent Neural Networks (RNNs). Reported error rates were 7.82\% and 6.93\% on the Yelp and Amazon benchmarks, respectively. Notably, on Yelp, the \textit{n}-grams model outperformed the CCNNs with a 6.36\% error rate. The study highlighted several influencing factors on model performance, including dataset size, text curation, and the alphabet used, specifically distinguishing between uppercase and lowercase letters.

In a separate study,~\cite{shen2017learning} introduced a unique approach using CNNs with meta-networks to learn context-sensitive convolutional filters for text processing. Applying this approach on Yelp, they achieved a lower error rate of 4.89\% compared to the former CCNNs approach. However, deeper networks pose increased computational complexity, impacting practical applications. Addressing this,~\cite{johnson2017deep} proposed shallow word-based Deep Pyramid CNNs (DPCNN) for text categorization. They studied deepening word-level CNNs to capture comprehensive text representations without significantly increasing computational costs. Evaluating on Yelp and Amazon datasets, their method achieved error rates of 7.88\% and 7.92\%, respectively.

\subsubsection{Deep Learning on Movie Reviews SA}
To commence,~\cite{shirani2014applications} explored the performances of RNNs and CNNs architectures for SA of movie reviews. They utilized pre-defined 300-dimensional vectors from word2vec~\cite{mikolov2013efficient} instead of training word vectors along with other parameters using samples. The study indicated that CNNs outperformed RNNs, achieving the best accuracy of 46.4\% on the SST dataset. It was concluded that basic RNNs were inefficient in capturing the structural and contextual properties of sentences. Basic RNNs encounter issues such as vanishing or exploding gradients, leading to model underfitting and overfitting when networks become very deep. Addressing this challenge,~\cite{rusch2020coupled} proposed Coupled Oscillatory RNN (CoRNN), a time-discretization of a system of second-order ordinary differential equations, which mitigated the exploding and vanishing gradient problem. CoRNN achieved 87.4\% accuracy on IMDb by precisely bounding the gradients of hidden states.

LSTMs also contribute to mitigating these problems.~\cite{bodapati2019sentiment} applied LSTM on movie review SA, exploring different hyperparameters like dropout, number of layers, and activation functions. Their LSTM configuration, including embedding, LSTM layer, dense layer, 0.5 dropout, and 100 LSTM units, achieved an accuracy of 88.46\% on IMDb. Although LSTMs handle longer sequences efficiently, whether the incorporated gates in the LSTM architecture offer sufficient generalization or additional data training is required remains unclear~\cite{siami2019performance}. 
To address this,~\cite{singh2020revisiting} applied a BiLSTM network using a mixed objective function for text classification, employing both supervised and unsupervised approaches. Their study showcased that a simple BiLSTM model using maximum likelihood training delivered competitive performance in polarity classification, reporting a 6.07\% error rate.

Although BiLSTM displayed superior results compared to other deep learning methods, room for performance enhancement remains at a 6.07\% error rate. Hence, in our work, BiLSTM  is adopted to further improve performance in this task.

\subsection{BERT}
This section delves into works that utilize BERT for SA and specifically for analyzing movie reviews.

\subsubsection{BERT and SA}
BERT stands out as a renowned SOTA language model for its exceptional performance across various NLP tasks. For instance, in the realm of SA,~\cite{xu2020understanding} delved into attention mechanisms and pre-trained hidden representations of BERT for Aspect-Based SA (ABSA). Their analysis revealed BERT's utilization of minimal self-attention heads to encode contextual words, such as prepositions or pronouns indicating an aspect, and opinion words associated with aspects. Conversely,~\cite{li2019exploiting} explored the potential of contextualized embeddings from BERT in an end-to-end ABSA task, focusing on integrating BERT embeddings with various neural models. Their findings showcased the impressive performance of BERT when combined with models like Gated Recurrent Units (GRU).

In a similar vein,~\cite{gao2019target} leveraged the pre-trained BERT model for target-dependent sentiment classification, examining its context-aware representation for potential improvements in ABSA. Their study revealed that coupling BERT with complex neural networks previously effective with embedding representations did not add substantial value to ABSA.

Other investigations of BERT involve transfer learning approaches.~\cite{sun2019utilizing} fine-tuned a pre-trained BERT model for ABSA by transforming ABSA into a sentence-pair classification task, achieving an impressive 92.8\% accuracy on the SentiHood dataset. Meanwhile,~\cite{xu2019bert} explored BERT for fine-tuning on Review Reading Comprehension (RRC) and ABSA tasks, generating a ReviewRC dataset from a benchmark for ABSA. Their novel post-training fine-tuning approach on BERT achieved an accuracy of 90.47\%.

These studies collectively showcase effective fine-tuning techniques with BERT, particularly in ABSA tasks. In our work, inspired by the coupling technique used in~\cite{li2019exploiting}, we opt to couple BERT with BiLSTM for the movie reviews SA task.

\subsubsection{BERT and Movie Reviews SA}
In the realm of movie reviews,~\cite{maltoudoglou2020bert} employed BERT to transform words into contextualized word embeddings. They fine-tuned BERT's parameters using the IMDb movie reviews corpus through Inductive Conformal Prediction (ICP), achieving an accuracy of 92.28\%. In contrast,~\cite{alaparthi2021bert} pursued a different approach, comparing BERT against SentiWordNet\cite{baccianella2010sentiwordne}, logistic regression, and LSTM for Movie Reviews SA on the IMDb dataset. Their study aimed to ascertain the relative efficacy of the four SA algorithms, highlighting the undeniable superiority of the pre-trained advanced supervised BERT model in text sentiment classification. BERT notably outperformed other models, achieving an accuracy of 92.31\%. Notably, both studies focused on binary classification.

Conversely,~\cite{munikar2019fine} employed BERT for both binary and fine-grained classifications on SST-2 and SST-5 datasets, respectively. Their model showcased superior performance compared to other deep learning-based models, such as CNN and RNN, achieving an accuracy of 93.7\% on SST-2 and 55.5\% on SST-5 tasks.

It is evident that deep learning techniques have proven to be the most accurate approaches for SA. Transfer learning, particularly fine-tuning BERT, consistently yields outstanding results. However, despite the reported advancements and the capabilities of BERT, there remains significant potential to further enhance the pre-trained language model's performance in this field. Furthermore, many studies have predominantly focused on either polarity sentiment classification or fine-grained sentiment classification, overlooking an exploration into how the categorical scope of sentiment polarities affects model robustness. Additionally, most researchers have not evaluated their approaches across a wide spectrum of available benchmark datasets for SA. For instance, results are often reported exclusively for IMDb or SST variants.

Therefore, our current work aims to fine-tune BERT by coupling it with BiLSTM for both polarity classification and fine-grained sentiment classification, given that these techniques have demonstrated superior performance in the existing literature. Leveraging transfer learning insights from~\cite{semwal2018practitioners}, we extend previous methodologies by computing the overall polarity of sentiments, as demonstrated in~\cite{arasteh2021will}. While a prior study computed overall polarity based on a single output vector derived from Twitter replies using LSTM for three-class classification, we intend to compute this using the output vector obtained from BERT coupled with BiLSTM, tailored to each specific classification task. Our experimental evaluations will encompass diverse benchmark datasets as shown in Section \ref{Experiments}.

\section{Methodology} \label{Methodology}
This section discusses various techniques employed in this study, encompassing SA, BERT, BiLSTM, fine-tuning BERT with BiLSTM, different classification tasks, techniques for performance enhancement, computation of overall polarity, and an overview of the entire study. 

\subsection{SA}
SA is a sub-domain of opinion mining, focusing on extracting emotions and opinions regarding a specific topic from structured, semi-structured, or unstructured textual data~\cite{gadekallu2019application}. It can be approached either as a binary or fine-grained sentiment classification task. In binary classification, machine learning models categorize text into positive or negative sentiment categories.
In contrast, fine-grained classification involves utilizing more than two sentiment classes, enabling the computation of multi-scale sentiment distribution. In our context, we investigate both classifications. Our aim is to explore the robustness and consistency of the model in generalization across varying categorical scopes of sentiment polarities.

\subsection{Model Architecture}
The model architecture consists of BERT, BiLSTM and a fully connected dense layer.
\subsubsection{BERT}
BERT was introduced by researchers from Google~\cite{devlin2018bert}. BERT primarily focuses on pre-training deep bidirectional representations from unlabeled text by jointly considering both left and right contexts across all layers of the model. Consequently, BERT can be fine-tuned with a single additional layer for various downstream tasks such as SA, question answering, and more. Pre-training of BERT involved two unsupervised tasks.

\paragraph{Masked Language Modeling} For this task, 15\% of the tokens within the input sequence undergo random masking. Subsequently, the complete input sequence is processed through a deep bidirectional transformer encoder, and the output softmax layer is designed to predict the masked words\cite{devlin2018bert}.

\paragraph{Next Sentence Prediction} BERT establishes a relationship between two input sentences, denoted as sentence \textit{A} and sentence \textit{B}, by predicting whether these sentences logically follow each other within a specific monolingual corpus. During training, 50\% of the inputs consist of sentence pairs where the second sentence is the immediate subsequent sentence in the source document. Conversely, the remaining 50\% involve pairs where a random sentence from the corpus is chosen as the second sentence \cite{munikar2019fine}.

BERT processes a sequence of input tokens simultaneously due to its multiple attention heads, and the model is primarily available in two sub-models: BERT\textsubscript{BASE} and BERT\textsubscript{LARGE}. In this work, we utilize BERT\textsubscript{BASE}, which comprises 12 layers, 768 hidden states, 12 attention heads, and 110M parameters. In contrast, BERT\textsubscript{LARGE} features approximately twice the specifications of BERT\textsubscript{BASE}. Specifically, the uncased version of BERT\textsubscript{BASE}, referred to as \textit{bert-base-uncased}, which processes input tokens in lowercase, is employed in this study.  

In BERT, there is a specific format for input tokens. The initial token of each sequence is labeled as [CLS]. This token corresponds to the final hidden layer, gathering and aggregating all the information within the input sequence, particularly for classification tasks. To distinguish between sentences within a single input sequence, two methods are employed: the use of a special token, [SEP], for separation and the addition of a learned embedding to each token, thereby identifying the sentence to which it belongs.

\begin{figure}[!h] 
\centering
\includegraphics[width=0.32\textwidth]{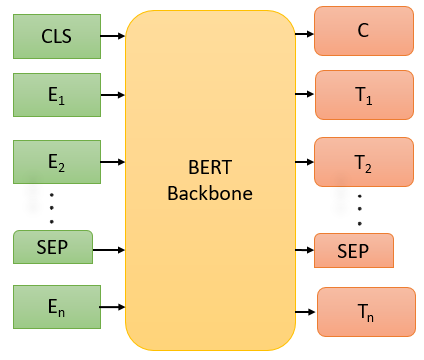}
\caption{Simplified diagram of BERT}
\label{fig:fig1}
\end{figure}

Figure~\ref{fig:fig1} illustrates a simplified diagram of BERT. $E_n$ denotes the input representation of a single token, generated by summing the respective token, segment, and position embeddings. The \textit{BERT Backbone} symbolizes the primary processing carried out by BERT, while $T_n$ represents the hidden state corresponding to token $E_n$. Additionally, $C$ represents the hidden state corresponding to the aggregated token [CLS]. Consequently, we utilize $C$ as the input for the fine-tuning components in sentiment classification.

\subsubsection{BiLSTM}
 BiLSTM is an LSTM variant that processes input features in both forward and backward directions~\cite{graves2005framewise}. This bidirectional characteristic provides BiLSTM an advantage in effectively capturing higher-level sentiment representations from the BERT hidden state \textit{C}. Additionally, BiLSTM inherits the advantageous features of LSTM, including the ability to retain long-distance temporal dependencies and avoiding the issues of vanishing or exploding gradients during backpropagation through time. The enhancement of the performance of the model by supplementing it with the BiLSTM module, as demonstrated in experiments, outperformed the usage of solely BERT with a dense layer.

\subsubsection{Dense Layer}
The hidden state from BiLSTM feeds into a dense layer. This layer, based on the BiLSTM output, generates a higher-level feature set that is readily distinguishable for the targeted number of classes. Finally, a softmax layer is added atop the dense layer to yield the predicted probability distribution for the target classes.

\subsection{Fine-tuning BERT with BiLSTM} 
Because BERT is pre-trained, there is no necessity to train the entire model from the beginning~\cite{devlin2018bert}. Consequently, we simply transfer knowledge from BERT to the added fine-tuning layer and train this layer for SA.  

In this study, the fine-tuning process operates as follows. After data pre-processing, three layers are established, one utilizing BERT and the subsequent layers involving BiLSTM and dense networks. The data pre-processing phase generates two input values, known as \textit{attention masks} and \textit{input ids}. These serve as the input embeddings to the model.

The input embeddings are then passed through the BERT module. The dimensionality of these embeddings is contingent on various factors, including the input sequence length, batch-size, and the number of units in BERT's hidden state.

Subsequently, the BiLSTM assimilates information from BERT and forwards it to the dense layer, which predicts the corresponding classes for the input features through the succeeding softmax layer. BERT and BiLSTM share the same hyperparameters, detailed in Section \ref{Experiments} under experimental settings.

\begin{figure*}[!t] 
\center
\includegraphics[width=0.9\textwidth]{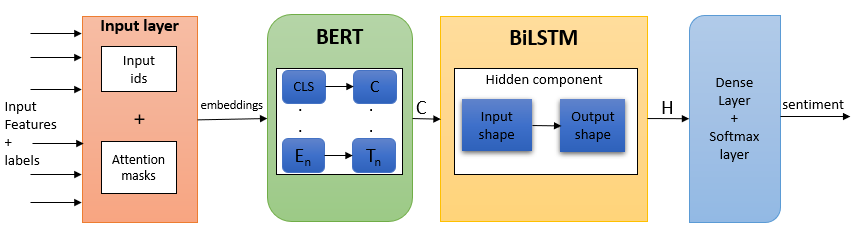}
\caption{Fine-tuning part of BERT with BiLSTM}
\label{fig:fig2}
\end{figure*}

The fine-tuning process is depicted in Figure \ref{fig:fig2}. Within the figure, \textit{Input features} represent tokens in a review, \textit{Input ids} symbolize an input sequence, and \textit{Attention masks} are binary tensors that highlight the position of the padded indices in a particular sequence to exclude them from attention. These masks use binary values 1 to indicate positions that require attention and 0 for padded values. Padding ensures uniform sequence lengths for input data sentences, a common practice in NLP. Consequently, the padded information is not considered part of the input and has minimal impact on the model's generalization.

The output from BERT matches the dimensionality of the input to BiLSTM, set at 768 and represented by \textit{C}. Only \textit{C} is transmitted to BiLSTM. The BiLSTM layer includes a single hidden component, followed by a dense layer that receives the hidden state \textit{H} from BiLSTM. 

\subsection{Classification}
In this study, BERT is fine-tuned for both binary (polarity) sentiment classification and fine-grained SA. The fine-grained classification aspect encompasses three distinct tasks: three-class (3-point scale), four-class (4-point scale), and five-class (5-point scale) classifications.

\subsubsection{Polarity classification} In this context, polarity classification entails classifying a movie review \textit{R} as either conveying a positive or a negative sentiment polarity. This fundamental task is often a cornerstone of SA, as it primarily focuses on discerning between positive and negative sentiments within a text~\cite{mesnil2014ensemble}.

Additionally, to assess the robustness of the model in handling varying categorical scopes of sentiment polarities, we further perform fine-grained sentiment classification. This involves utilizing different classification scales to gauge how the polarity of an individual review and the overall polarity of a collection of reviews change as the classification becomes more detailed. For instance, if a review is classified as negative in a binary classification dataset, we aim to understand how this polarity aligns when labels are expanded to include highly negative, negative, neutral, positive, and highly positive categories. We extend this analysis to overall sentiment polarity as well. Consequently, we conduct the following fine-grained classification tasks.

\subsubsection{Three-class classification} 
This expands upon binary classification by introducing a neutral class to account for instances where reviewers might not distinctly assign a positive or negative sentiment to a movie due to ambiguous sentiments or a lack of clear preference~\cite{arasteh2021will}. Sometimes, the review might contain a balance of positive and negative words, leading to ambiguity in sentiment. Hence, a neutral polarity is introduced to accommodate such cases. In the context of three-class classification, the task is defined as follows: given a movie review \textit{R}, classify it as carrying a negative, neutral, or positive sentiment polarity.

Alternatively, the output vector from binary classification can be directly adapted into three-class classification without altering the label scope in the training data or restarting the training process for the three-class task. By employing a sigmoid activation function or a softmax layer that assigns varying confidence levels to negative and positive classes, a neutral class can be incorporated using a delta value, $\delta$. In this approach, the actual output label can be replaced by a neutral label if the discrepancy between the probabilities of the original two classes is smaller than $\delta$. However, this method necessitates a careful definition of $\delta$ to ensure meaningful and accurate outcomes. 

\subsubsection{Four-class classification} 
\begin{figure}[!h]
\centerline{\includegraphics{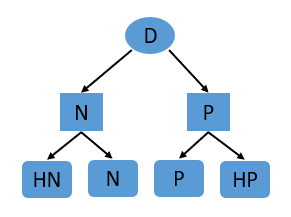}}
\caption{Binary Tree Splitting}
\label{fig:fig3}
\end{figure}

This task specifically targets the IMDb dataset. To extend the binary IMDb classification to four classes, we adopt a hierarchical approach using binary tree splitting. This technique, initially introduced in~\cite{mola1992two}, leverages binary segmentation to identify homogeneous nodes within a tree structure. In our scenario, negative reviews are divided into highly negative and negative, while positive reviews are categorized into positive and highly positive, as illustrated in Figure \ref{fig:fig3}. Here, \textit{D} represents the dataset, while \textit{N}, \textit{P}, \textit{HN}, and \textit{HP} symbolize Negative, Positive, Highly Negative, and Highly Positive reviews, respectively. A detailed explanation of the binary tree splitting method's application is provided in Section \ref{Experiments} within the IMDb dataset analysis. Consequently, the four-class classification is defined as follows: given a movie review \textit{R}, classify it based on whether it carries a highly negative, negative, positive, or highly positive sentiment polarity.

\subsubsection{Five-class classification} Five classes are used here as in \cite{munikar2019fine} and \cite{rosenthal2017semeval}. While  \cite{rosenthal2017semeval} used the output vector obtained from this classification to estimate the distribution of data examples across the five classes, we use the vector to find the overall sentiment polarity of all the predictions. Five-class task is defined as follows. Given a movie review \textit{R}, classify it as whether carrying a highly negative, negative, neutral, positive, or highly positive sentiment polarity.

\subsection{Accuracy Improvement Approaches}
We utilized distinct data oversampling and augmentation techniques individually to improve the accuracy  of the model for fine-grained sentiment classification.

\subsubsection{Oversampling} The Synthetic Minority Over-sampling TEchnique (SMOTE) was introduced in~\cite{chawla2002smote} to address imbalanced datasets problem and enhance model performance. As SMOTE primarily operates with numerical input data, we initially converted reviews from the minority classes into their corresponding numerical features. Subsequently, we used these features as input for SMOTE, which conducted oversampling to balance the dataset. 

\subsubsection{Augmentation}Data augmentation serves to address class imbalance and maximize information extraction from limited resources \cite{rizos2019augment}. We employed NLP Augmenter (NLPAUG), a technique leveraging operations based on abstractive summarization and synonym replacement driven by the proximity of word embedding vectors. Our experiments found this technique successful, and we depict it in Eq. \eqref{Eq1}:
\begin{equation} \label{Eq1}
V\textsubscript{AUG} = F(V\textsubscript{IN})
\end{equation}

Here, $V_{\text{AUG}}$ represents the output matrix of augmented sentences, where $F$ symbolizes the abstractive summarization augmentation function applied to the input matrix $V_{\text{IN}}$ containing raw text data. Both matrices contain $n$ vectors, corresponding to the misrepresented (minority) classes within the dataset. Furthermore, the input vectors comprise randomly sampled data examples from specific misrepresented classes, establishing a one-to-one mapping between input and output vectors.

After applying these techniques, we initially combined the output from SMOTE with the original input data and proceeded to train the model using the SST-5 benchmark dataset. Subsequently, we performed a similar procedure using \textit{V\textsubscript{AUG}}.

\subsection{Overall Sentiment Polarity}
 We define the overall sentiment polarity as follows: Given an output vector \textit{V} from BERT+BiLSTM containing sentiment labels for \textit{N} reviews, we compute the dominant sentiment polarity within the vector. To derive the overall sentiment polarity of the reviews, we input the output vector of BERT+BiLSTM into a heuristic algorithm. In this process, BERT+BiLSTM initially predicts the sentiment category for each review, aggregating the results in an output vector. The occurrences of each class label within the output vector are tallied, and the result is passed to the heuristic algorithm to determine the predominant sentiment polarity for all the reviews collectively.

The algorithm has been modified to compute the overall sentiment polarity from the output vectors of binary, four-class, and five-class classifications. The functionality of the algorithm is dependent on the number of classes in the output vector, necessitating the derivation of three variants of the algorithm for these different classification tasks.

Algorithm \ref{alg2} outlines the computation of the overall polarity from the output vector of the three-class classification. Initially, if the proportion of neutral reviews exceeds a threshold, set at $85\%$, the overall polarity is designated as neutral. This threshold acknowledges that a majority of reviews might tend towards a neutral sentiment rather than distinctly positive or negative.

Next, the algorithm considers the distribution of negative and positive sentiments in the output vector. Typically, individual reviews seldom express exclusively positive or negative sentiments. Therefore, if the number of negative reviews exceeds positive reviews by at least 1.5 times, or vice versa, the dominant sentiment determines the overall polarity. This criterion ensures that a sentiment must significantly outweigh the other to influence the overall polarity.

Finally, when the counts of positive and negative reviews are nearly equal, indicating a lack of a clear dominance between positive and negative sentiments, the overall polarity is categorized as neutral once more. This scenario implies a balanced representation of both sentiments across the reviews.
 
To compute the overall polarity from the binary classification task, Algorithm \ref{alg1} is derived from Algorithm \ref{alg2} with modifications to account for the absence of a neutral polarity in binary classification. While Algorithm \ref{alg2} considers the presence of a neutral sentiment, binary classification does not include this category. Therefore, in Algorithm \ref{alg1}, the logic is adjusted to directly assess the quantities of positive and negative reviews. However, in cases where neither category dominates the sentiment output (i.e., quantities of positive and negative reviews are approximately similar), a neutral sentiment polarity is introduced for the overall polarity computation, representing a tie between positive and negative sentiments.

 In our formulations, we employed variable coefficients, namely 1.2 and 1.5, to ascertain the majority gap for decision-making within the algorithms, for comparisons within the algorithms.

\begin{algorithm}[t!] \label{alg2}
\caption{Overall sentiment polarity computation from three-class classification output vector.}
\KwResult{Dominating sentiment polarity for all reviews.} 

\uIf{$\#$total neutral reviews $> 85\%$ of the total reviews}{\textit{overall polarity} $\gets$ \textit{neutral}\;}
\uElse{
    \uIf{$\#$total positive reviews $> $1.5 $\times$ \# of total negative reviews}{\textit{overall polarity} $\gets$ \textit{positive}\;}
    
    \uElseIf{$\#$total negative reviews $> $1.5 $\times$ \# of total positive reviews}{\textit{overall polarity} $\gets$ \textit{negative}\;}
    \uElse{\textit{overall polarity} $\gets$ \textit{neutral}\;}
}
\end{algorithm}

\DontPrintSemicolon
%second algorithm
\begin{algorithm} [t!] \label{alg1}
\caption{ Overall sentiment polarity computation from binary classification output vector.}
\KwResult{Dominating sentiment polarity for all reviews.} 

\uIf{$\#$total positive reviews $> $1.2 $\times$ \# of total negative reviews}{\textit{overall polarity} $\gets$ \textit{positive}\;}

\uElseIf{$\#$total negative reviews $> $1.2 $\times$ \# of total positive reviews}{\textit{overall polarity} $\gets$ \textit{negative}\;}

\uElse{\textit{overall polarity} $\gets$ \textit{neutral}\;}
\end{algorithm}

%third algorithm
\begin{algorithm} \label{alg3}
\caption{ Overall sentiment polarity computation from four-class classification output vector.}
\KwResult{Dominating sentiment polarity for all reviews.} 

\uIf{$\#$ (highly negative reviews + negative reviews) $> $1.2 $\times $ \# of (positive reviews + highly positive reviews)}
{
     \uIf{$\#$highly negative reviews $> $1.5 $\times$ \# of negative reviews}{\textit{overall polarity} $\gets$ \textit{highly negative}\;}
     \uElse{\textit{overall polarity} $\gets$ \textit{negative}\;}
}

\uElseIf{$\#$(positive reviews + highly positive reviews) $> $1.2 $\times$ \# of (highly negative reviews + negative reviews)}
{
     \uIf{$\#$highly positive reviews $> $1.5 $\times$ \# of positive reviews}{\textit{overall polarity} $\gets$\textit{ highly positive}\;}
     \uElse{\textit{overall polarity} $\gets$ \textit{positive}\;}
}
\uElse{\textit{overall polarity} $\gets$ \textit{neutral}\;}
\end{algorithm}

Algorithm \ref{alg1} serves as the foundation for Algorithm \ref{alg3}, enabling the computation of overall polarity from the four-class output vector of BERT+BiLSTM. This algorithm operates hierarchically, taking into account the binary tree splitting illustrated in Figure \ref{fig:fig3}. The hierarchical comparison starts with base classes, progressing to sub-classes within a base class that holds the majority of samples. The process involves aggregating the sample counts of all fine-grained classes under each super class. For instance, the total for the negative super class is derived from the highly negative and negative sub-class totals.

Three distinct scenarios are considered within the algorithm. In the first case, if the total number of negative reviews across super classes exceeds the total number of positive reviews by at least 1.2 times, and the count of highly negative sub-class reviews is at least 1.5 times that of the negative sub-class, a highly negative overall polarity is assigned. Conversely, if the positive reviews surpass the negative reviews by 1.2 times, and the highly positive sub-class reviews dominate by 1.5 times over the positive sub-class, an overall positive polarity is assigned. Finally, if no significant dominance is observed between the total numbers of negative and positive reviews within the base classes, a neutral overall polarity is assigned.

For base classes, the threshold of 1.2 often works best to determine the dominating class. However, in the more finely-grained sub-classes, such as highly positive and highly negative, a threshold of 1.5 tends to perform better. This distinction is due to the increased granularity of the sub-classes, requiring a higher sample size of the dominating subclass to designate its label as the overall polarity.

%fourth algorithm
\begin{algorithm} \label{alg4}
\caption{Overall sentiment polarity computation from five-class classification output vector.}
\KwResult{Dominating sentiment polarity for all reviews.} 
\uIf{$\#$total neutral reviews $> 85\%$ of the total reviews}{\textit{overall polarity} $\gets$ \textit{neutral}\;}
\uElse{
    \uIf{($\#$highly negative reviews + $\#$negative reviews) $> $1.2 $\times$ \# of (positive reviews + highly positive reviews)}
    {
         \uIf{$\#$highly negative reviews $> \#$1.5 $\times$ negative reviews}{\textit{overall polarity} $\gets$ \textit{highly negative}\;}
         \uElse{\textit{overall polarity} $\gets$ \textit{negative}\;}
    }
    
    \uElseIf{$\#$(positive reviews + highly positive reviews) $> $1.2 $\times$ \# of (highly negative reviews + negative reviews)}
    {
         \uIf{$\#$high positive reviews $> $1.5 $\times$  \# of positive reviews}{\textit{overall polarity} $\gets$ \textit{highly positive}\;}
         \uElse{\textit{overall polarity} $\gets$ \textit{positive}\;}
    }
    \uElse{\textit{overall polarity} $\gets$ \textit{neutral}\;}
    }
\end{algorithm}

For five-class classification, Algorithm \ref{alg4} extends from Algorithm \ref{alg3} with an additional initial step. Initially, the overall polarity is evaluated as neutral if the proportion of neutral reviews exceeds a designated threshold, typically set at 85\%. Following this initial consideration, the subsequent steps in Algorithm 3 are maintained without alteration.

A simplistic method for determining overall polarity involves tallying the count of labels for each class within the BERT+BiLSTM output vector and assigning overall polarity based on the majority class. However, this basic approach may not adequately represent the dominant sentiment. Consider binary classification, if the output vector consists of 49 positive reviews and 51 negative reviews, the overall polarity will be negative. Yet, a difference of 2 is not conclusive enough to determine the prevailing sentiment across all reviews.

Furthermore, the levels of positivity or negativity vary among reviews in the datasets. To address this, we employ specific formulations in the respective algorithms. These formulations require a significantly higher majority within a class in the output vector to assign its label as the overall sentiment polarity; otherwise, the overall sentiment polarity defaults to neutral.

 Additionally, we refined the coefficients to 1.2 and 1.5 in the algorithms based on various observations of the model's behavior across different scenarios. These empirically determined values significantly improved the accuracy of overall polarity computation, closely reflecting the original sentiment polarity of input features.

\subsection{Overview of the  Work}

\begin{figure*}[!t]
\center
\includegraphics[width=0.85\textwidth]{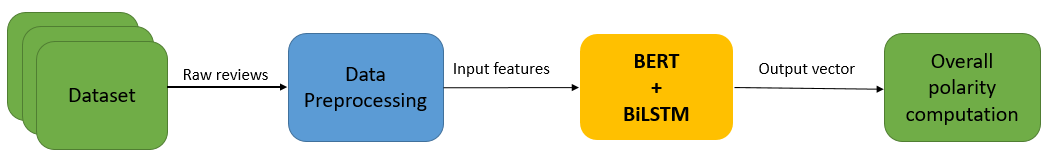}
\caption{Overview of our work} 
\label{fig:fig4}
\end{figure*}

Illustrated in Figure \ref{fig:fig4}, the overview of our work encapsulates several key steps. Initially, raw text data undergoes preprocessing to transform it into features compatible with BERT. Subsequently, these features are fed into BERT+BiLSTM through the fine-tuning layer, which configures the necessary hyperparameters for BERT+BiLSTM. Finally, the output vector from BERT+BiLSTM predictions is utilized to compute the overall sentiment polarity. 

\section{Experiments} \label{Experiments}
This section provides an insight into the datasets utilized in this study, followed by an in-depth exploration of the data pre-processing techniques. Subsequently, it delves into the experimental settings, elucidates the evaluation metrics employed in the experiments, and culminates in a comprehensive discussion of the experimental findings.

\subsection{Datasets}
The datasets employed within this study revolve around movie reviews meticulously annotated for SA across a diverse array of scales, specifically, 2-point, 3-point, 4-point, and 5-point gradations. The subsequent elucidation encapsulates the detailed descriptions of these datasets.

\subsubsection{IMDb movie reviews} 
The IMDb movie reviews dataset~\cite{maas2011learning}, stands as a widely embraced binary SA collection, encompassing a colossal 50,000 reviews sourced from the Internet Movie Database (IMDb). This assortment impeccably balances negativity and positivity, presenting an equal split between negative and positive reviews. Within this study, the dataset assumes a pivotal role in binary classification tasks, further expanding its utility to encompass three-class and four-class classification endeavors. Comprising three fundamental columns (reviews, sentiment score, and label), the dataset's sentiment scores range from integers 1 to 10, excluding the neutral ground of 5 and 6. Reviews carrying scores between 1 to 4 are distinctly labeled as negative, while those within the 7 to 10 spectrum are unequivocally regarded as positive.

In extending the dataset towards a four-class classification paradigm, our approach leveraged the concept of binary-tree splitting, visually represented in Figure \ref{fig:fig3}. Each review within the dataset boasts eight distinct sentiment scores, owing to the absence of scores 5 and 6. Our classification strategy involves categorizing reviews bearing scores 1 and 2 as highly negative, while scores 3 and 4 are tagged as negative. Conversely, scores 7 and 8 merit a positive label, and ratings of 9 and 10 earn the designation of highly positive. For the three-class model, our breakdown was as follows: scores 1, 2, and 3 corresponded to negative sentiment, scores 4 and 7 indicated a neutral stance, and ratings of 8, 9, and 10 were indicative of positive sentiment. We refer to these modified datasets as IMDb-2, IMDb-3, and IMDb-4, representing the binary, three-class, and four-class renditions of the original dataset, respectively. 

\subsubsection{SST} The Stanford Sentiment Treebank (SST) stands as a punctiliously curated corpus adorned with fully labeled parse trees, offering a profound exploration into the intricate nuances of sentiment's compositional impact on language. Originating from the dataset introduced in \cite{socher2013recursive}, SST comprises a collection of 11,855 individual sentences meticulously extracted from movie reviews. These sentences underwent parsing via the Stanford parser, resulting in an extensive repository of 215,154 unique phrases diligently annotated by three human judges. Each phrase within this dataset bears a label denoting its sentiment, falling within the spectrum of negative, somewhat negative, neutral, somewhat positive, or positive, equating to the highly negative, negative, neutral, positive, and highly positive labels used within our annotations.

The SST dataset manifests in two distinct versions: SST-5, known as SST fine-grained, employs five labels to characterize sentiment nuances, while SST-2, termed SST binary, simplifies the classification into two primary labels, negative and positive. Within SST-2, negative sentiments encompass judgments labeled as negative or somewhat negative, while positive sentiments entail those marked as somewhat positive or positive. Notably, neutral reviews find no place within SST-2. In our present study, we harness SST-2 for binary classification tasks and SST-5 for more intricate five-class classification endeavors, thus delving deeper into the subtleties of SA within the corpus.

\subsubsection{MR Movie Reviews} The MR movie reviews dataset encompasses a wealth of movie review documents carefully labeled based on their overarching sentiment polarity, whether they lean positively or negatively, or subjective rating, such as nuanced assessments like two and a half stars. Additionally, it contains sentences categorized according to their subjectivity status, distinguishing between subjective or objective content, and polarity.

Within the scope of this paper, we specifically employ the version introduced in \cite{pang2005seeing}, a rigorously curated subset comprising 5,331 positive and 5,331 negative processed reviews. In our experiments, we exclusively harness the MR movie reviews dataset for a singular purpose: the binary classification task. This dataset serves as the cornerstone of our experiments, focusing solely on the polarity aspect of SA.

\subsubsection{Amazon Product Data dataset} This dataset is an expansive repository housing product reviews and metadata sourced from Amazon, encompassing a staggering 142.8 million reviews dating from May 1996 to July 2014. Its comprehensive scope spans reviews, product metadata, and interlinkages. Initially introduced in \cite{fang2015sentiment} as a resource for SA utilizing product review data, it was further utilized in \cite{he2016ups} to construct a recommender system operating within a collaborative filtering framework tailored specifically for Amazon products.

Within the context of our study, our focal point centers solely on video reviews within this extensive dataset. Initially, the dataset featured labels graded from scores 1 to 5, signifying a spectrum of polarity strength ranging from highly negative to highly positive sentiments. To streamline our analysis toward binary classification, we undertook a transformation: scores 1 and 2 were consolidated into a negative label, while scores 4 and 5 were amalgamated into a positive label. Simultaneously, score 3, representing a neutral stance, was omitted, akin to the approach adopted in SST-2 \cite{socher2013recursive}.

For the sake of clarity and specificity within our research, we distinguish between Amazon-2 and Amazon-5, representing the binary and five-class versions of the dataset, respectively. Table \ref{tab0} provides a comprehensive statistical overview of the datasets employed in our experiments, where categorical labels like H POS (Highly Positive), POS (Positive), NEG (Negative), and H NEG (Highly Negative) elucidate distinct sentiment categories.

\begin{table*}[!t]
\caption{Statistics of the datasets divided into training and test sets }
\begin{center}
\begin{tabular}{|l|r|r|r|r|r|r|r|r|r|r|}
\hline
\textbf{Dataset}&\multicolumn{5}{|c|}{\textbf{Train samples}} &\multicolumn{5}{|c|}{\textbf{Test samples}}\\
\cline{2-11} 
\textbf{} & H POS &POS &NEUTRAL & NEG &H NEG 
& H POS &POS &NEUTRAL & NEG &H NEG  \\
\hline
IMDb-2 &- &12500 &- &12500 &- &- &12500 &- &12500 &-  \\
\hline
MR &- &4264 &- &4265 &- &- &1067 &- &1066 &-  \\
\hline
SST-2  &-  &4300  &- &4244 &- &- &886 &- &1116 &-  \\
\hline
Amazon-2 &-  &239660 &- &37056 &- &- &59949 &- &9231 &-  \\
\hline
IMDb-3 &-  &18227 &4816 &14958 &- &- &4556 &1204 &3739 &-  \\
\hline
IMDb-4 &11471  &8530 &- &8234 &11767 &2867 &2132 &- &2058 &2941  \\
\hline
SST-5 &1482  &2489 &1794 &2512 &1208 &370 &622 &448 &628 &302\\
\hline
Amazon-5 &182000  &57688 &27767 &15168 &21863 &45500 &14421 &6941 &3791 &5465\\
\hline
\end{tabular}
\label{tab0}
\end{center}
\end{table*}

\subsection{Data pre-processing}
The data pre-processing step was commenced by eliminating empty reviews across all datasets, ensuring a clean and consistent starting point for our analysis. Subsequently, the focus shifted towards implementing recommended data pre-processing measures aimed at translating the raw input data into a format compatible with BERT's comprehension. This undertaking encompassed two pivotal steps integral to the transformation process. 

Initially, we generate input examples utilizing the BERT-provided constructor, which operates based on three key parameters: \textit{text\_a}, \textit{text\_b}, and \textit{label}. Here, \textit{text\_a} encapsulates the body of text we aim to classify, specifically, the collection of movie reviews sans their associated labels. The \textit{text\_b} parameter, primarily utilized in scenarios involving sentence relationships such as translation or question answering, remains intentionally left blank given its minimal relevance to our current research focus. Meanwhile, \textit{label} serves as the container for the sentiment polarity labels associated with each movie review, representing an essential component of the input features.

For a deeper understanding of this foundational step, additional insights can be gleaned from the original BERT paper \cite{devlin2018bert}. Subsequently, we progress through a series of text pre-processing steps integral to our methodology.

\begin{itemize}
    \item Lowercase all text, since the lowercase version of BERT\textsubscript{BASE} is being used.
    \item Tokenize all sentences in the reviews. For example, ``this is a very fantastic movie", to `this', `is', `a', `very', `fantastic', `movie'.
    \item Break words into word pieces. That is  `interesting' to `interest' and `\#\#ing'.
    \item Map words to indexes using a vocab file that is provided by BERT.
    \item Add special tokens: [CLS] and [SEP], which are  used for aggregating information of the entire review through the model and separating sentences, respectively.
    \item Append index and segment tokens to each input to track a sentence that a specific token belongs to.
\end{itemize}

Following these processes, the tokenizer yields crucial outputs: \textit{input ids} and \textit{attention masks}. These outputs serve as pivotal components, subsequently utilized as inputs alongside the review labels for our model. These elements collectively form the foundational input data fed into our analysis framework.

\subsection{Experimental settings} \label{exp_settings}
We conducted fine-tuning on \textit{bert-base-uncased}, a variant of BERT designed to process lowercase tokens. While initializing most of BERT's layers from the model checkpoint, some layers required new initialization. Throughout the training process, we deliberately froze the initial layers containing BERT weights, allowing the primary focus to rest on the latter layers housing the fine-tuning components. These latter layers, holding trainable weights, were continuously updated to minimize loss during the training phase, specifically tailored to the downstream task of SA.

Numerous simulations were executed across the datasets to optimize the hyperparameters of the model. From our exhaustive experiments, the most optimal results emerged with specific hyperparameters. For binary classification, the model exhibited exceptional performance using the Adam optimizer~\cite{kingma2014adam}, a batch size of 32, a learning rate of 3e-5, an epsilon value of 1e-08, a maximum sequence length of 128, and employing binary cross-entropy loss. Training was conducted over 10 epochs, iterating through each batch.

Conversely, in all fine-grained classifications, a batch size of 64, a learning rate of 1e-4, a maximum sequence length of 256, a decay of 1e-5, alongside the Adam optimizer and sparse categorical cross-entropy loss, facilitated optimal outcomes. This specific version underwent training for 15 epochs, repeating steps for batches. Notably, we observed instances of overfitting when attempting to increase the number of epochs for these respective models.

To facilitate broader adoption and enable replication of our work, the code for this project is readily available \cite{bcode}, ensuring accessibility for interested parties.

\subsection{Evaluation Metrics}
In line with the evaluation metrics adopted in~\cite{munikar2019fine}, our approach also employs accuracy as a performance measure to assess the efficacy of our model in comparison to other models. Accuracy, in its essence, is straightforwardly defined as:

\begin{equation}
accuracy = \frac{\text{\textit{number of correct predictions}}}{\text{\textit{total number of predictions}}}\times 100
\end{equation}

\subsection{Results and Discussion} \label{results}
We delineate accuracy comparisons across various datasets and classification tasks in our study. Table \ref{tab1} illustrates binary classification results, encompassing a comparison between our model and others. Subsequently, in Table \ref{tab3}, we present outcomes for three-class and four-class classifications, exclusively on the IMDb dataset. Last, Table \ref{tab2} delves into five-class classification on SST-5 and Amazon-5 datasets. Across these tables, our model consistently surpasses all others in performance across diverse classification tasks on every dataset, solidifying its position as the new benchmark with SOTA accuracy levels.

The analysis of our results reveals a distinct trend: our model excels notably in binary classification tasks. However, a discernible pattern emerges as we move towards finer classification scales, evident across the IMDb, SST, and Amazon datasets showcased in Table \ref{tab1}, Table \ref{tab3}, and Table \ref{tab2}. As the classification granularity increases, the accuracy of our model experiences a consistent decline.

This decline in accuracy can be attributed to several factors. The primary reason lies in the heightened complexity that arises with an augmented number of classes within a dataset. As the distinctions between classes become more intricate, the task of accurate classification becomes increasingly challenging for the model.

Another contributing factor is the distribution of samples across these classes. Despite the proliferation of classes, the number of samples per class remains constrained by the original dataset size. Consequently, with the expansion of classes, each sentiment category possesses a reduced number of examples available for effective model learning. This scarcity of samples within each sentiment category hampers the ability of the model to learn distinct patterns for accurate predictions, particularly as the categorical spectrum of sentiment polarities widens within a dataset.

Through our experimental observations, a striking finding emerges: the model exhibits robustness and consistency, showcasing resilience even amidst alterations in the categorical scope of sentiment polarities within a specific dataset. To illustrate, we delve into a review snippet sourced from the IMDb benchmark. \textit{``This show is great. Not only is `Haruhai Suzumiya' a very well written anime show, it also reflects things like Philosophy, Science Fiction and a little religion. It\'s hilarious at some points and 'cute' (for lack of a better term) at others. Actually this may be effect to my lack of experience with Japanese anime shows, but it is one of the best of its genre I have seen. I mainly have to give credit to the writers. I haven\'t seen such brilliant scopes of imagination in a television show since the original Star Trek. I hope the writers continue to add strange new characters and give more insight on the already great characters that have been added"}. In the context of binary classification within IMDb-2, the model confidently designates the text as \textit{Positive}. Similarly, in the three-class classification of IMDb-3, it maintains this classification. However, when exploring the four-class classification in IMDb-4, the model assigns a \textit{Highly Positive} label. This classification aligns impeccably with the sentiment score embedded within the review text, a rating of 9 on a scale of 1 to 10 for goodness.

This astute adaptability demonstrates the proficiency of the model in discerning and leveraging nuanced semantic cues embedded within a dataset. It dynamically adjusts its classification strategy based on the chosen categorical scope, effectively capturing and reflecting the varying degrees of sentiment granularity present within the dataset.

\begin{table*}[ht]
\caption{Accuracy (\%) Comparisons  of Models on Benchmark Datasets for Binary Classification}
\centering
\begin{tabular}{|l|c|c|c|c|}
\hline
\textbf{Model name}&\multicolumn{4}{|c|}{\textbf{Dataset}} \\
\cline{2-5} 
\textbf{} & \textbf{\textit{IMDb-2}}& \textbf{\textit{MR}}& \textbf{\textit{SST-2}} & \textbf{\textit{Amazon-2}} \\
\hline
RNN-Capsule \cite{wang2018sentiment} &84.12& 83.80&82.77&82.68 \\
\hline
coRNN \cite{semwal2018practitioners} & 87.4 & 87.11 & 88.97&89.32\\
\hline
TL-CNN \cite{semwal2018practitioners} & 87.70 & 81.5 & 87.70&88.12\\
\hline
Modified LMU \cite{chilkuri2021parallelizing} & 93.20&93.15&93.10&93.67\\
\hline
DualCL \cite{chen2022dual} &- & 94.31& 94.91 &94.98\\
\hline
L Mixed \cite{sachan2019revisiting} &95.68 &95.72&-&95.81 \\
\hline
EFL \cite{wang2021entailment} & 96.10 & 96.90 & 96.90 &96.91\\
\hline
NB-weighted-BON+dv-cosine \cite{thongtan2019sentiment} & 97.40&-&96.55&97.55\\
\hline
SMART-RoBERTa Large \cite{jiang2019smart} & 96.34& 97.5 &96.61&-\\
\hline
\textbf{Ours }& \textbf{97.67}&\textbf{97.88} & \textbf{97.62} & \textbf{98.76}\\
\hline
\multicolumn{5}{l}{}
\label{tab1}
\end{tabular}
\end{table*}
%\Blindtext\Blindtext

%Table with results for fine-grained classification
\begin{table}[!t]
\caption{Accuracy (\%) Comparisons for Three and Four Class Classification on IMDd}
\begin{center}
\begin{tabular}{|l|c|c|c|c|}
\hline
\textbf{Model name}&\multicolumn{2}{|c|}{\textbf{Dataset}} \\
\cline{2-3}  & \textbf{\textit{IMDb-3}} & \textbf{\textit{IMDb-4}} \\
\hline
CNN-RNF-LSTM \cite{yang2018convolutional}& 73.71 & 63.78\\
\hline
DPCNN \cite{johnson2017deep} & 76.24 & 66.17 \\
\hline
BERT-large \cite{munikar2019fine} & 77.21 & 66.87 \\
\hline
\textbf{Ours} & \textbf{81.87} & \textbf{70.75} \\
\hline
\end{tabular}
\label{tab3}
\end{center}
\end{table}

%Table with results for fine-grained classification
\begin{table}[!t]
\caption{Accuracy (\%) Comporisons of Models on Benchmark Datasets for Five Class Classification}
\begin{center}
\begin{tabular}{|l|c|c|c|c|}
\hline
\textbf{Model name}&\multicolumn{2}{|c|}{\textbf{Dataset}} \\
\cline{2-3} & \textbf{\textit{SST-5}} & \textbf{\textit{Amazon-5}} \\
\hline
CNN+word2vec \cite{shirani2014applications} & 46.4 &48.85\\
\hline
TL-CNN \cite{semwal2018practitioners} & 47.2 & 58.1\\
\hline
DRNN \cite{wang2018disconnected}& - & 64.43\\
\hline
BERT-large \cite{munikar2019fine} & 55.5 & 65.83 \\
\hline
BCN+Suffix+BiLSTM-Tied+Cove \cite{brahma2018improved} & 56.2 & 65.92\\
\hline
RoBERTa+large+Self-explaining \cite{sun2020self} & 59.10 & -\\
\hline
\textbf{Ours} & \textbf{60.48} &  \textbf{69.68} \\
\hline
\end{tabular}
\label{tab2}
\end{center}
\end{table}

Our experimental findings, showcased in Table \ref{tab4} using the SST-5 dataset, include results concerning oversampling with SMOTE and augmentation using NPLPAUG. Two distinct approaches are highlighted: BERT+BiLSTM+SMOTE, utilizing SMOTE in conjunction with BERT and BiLSTM for SA, and BERT+BiLSTM+NLPAUG, employing NLPAUG for augmentation.

Surprisingly, the inclusion of SMOTE in our model exhibits no discernible impact on accuracy. Remarkably, the accuracy of a simpler model, BERT+BiLSTM, employing solely BERT and BiLSTM without any accuracy enhancement techniques, surpasses that of BERT+BiLSTM+SMOTE. This implies that the oversampling technique fails to impart semantic understanding to the model from the augmented data it produces.

Conversely, BERT+BiLSTM+NLPAUG showcases a performance boost, elevating the accuracy of the model from 58.44\% to 60.48\%. The rationale behind these observations lies in the input transformation process. SMOTE operates on text inputs that have been transformed into BERT features. This transformation introduces noise and disrupts the efficacy of SMOTE, as some semantic nuances are lost during this process. In contrast, NLPAUG operates directly on raw text data, facilitating an easier extraction of semantic information during the learning process. This direct access to raw text enables NLPAUG to enhance the performance of the model by leveraging the inherent semantics present in the data.

%SST-5 dataset results with Augmentation (NLPAUG) and oversampling (SMOTE).
\begin{table}[!t]
\caption{Accuracy (\%) of Our Model with Accuracy Improvement Techniques on  SST-5}
\begin{center}
\begin{tabular}{|l|c|}
\hline
\textbf{Classification task}  & \textbf{Accuracy} \\
\hline
BERT+BiLSTM & 58.44\\
\hline
BERT+BiLSTM+SMOTE & 58.36\\
\hline
\textbf{BERT+BiLSTM+NLPAUG}  & \textbf{60.48}\\
\hline
\end{tabular}
\label{tab4}
\end{center}
\end{table}

To encapsulate the discussion of results, we focus on the comprehensive computation of overall sentiment polarity across all datasets facilitated by our model, as detailed in Table \ref{tab5}. Here, \textit{OP} signifies Overall Polarity. We distinguish between \textit{Original OP}, known prior to input embeddings entering BERT+BiLSTM, and \textit{Computed OP}, derived after the model predicts sentiments for reviews.

Remarkably, the table demonstrates an intriguing consistency: \textit{Computed OP} aligns precisely with \textit{Original OP} across all datasets.\textit{ Original OP} was initially computed by tallying the sample counts for each label within the input features and then applying a specific heuristic algorithm tailored to the classification task. This initial computation was utilized as a benchmark to verify the accuracy of \textit{Computed OP}.

Furthermore, the table also highlights a consistent \textit{Computed OP} across different versions of the same dataset. This uniformity across dataset variations reinforces our confidence in the ability of the heuristic algorithm to accurately compute the expected overall polarity from the BERT+BiLSTM output vector. This confidence persists despite variations in the categorical scope of sentiment polarities within the movie reviews across different dataset iterations. This robustness ensures that the heuristic algorithm consistently delivers expected and accurate overall polarity computations, emphasizing its reliability and generalizability across diverse datasets.

%Overall polarity table
\begin{table}[t!]  
\caption{Overall Polarity Computation on All the Datasets}
\begin{center}
\begin{tabular}{|l|c|c|}
\hline
\textbf{Dataset}  & \textbf{Original OP}  & \textbf{Computed OP}$^{\mathrm{b}}$\\
\hline
IMDb-2 & Neutral & Neutral\\
\hline
IMDb-3 & Neutral & Neutral\\
\hline
IMDb-4 &  Neutral & Neutral\\
\hline
MR reviews & Neutral & Neutral\\
\hline
SST-2 & Neutral & Neutral\\
\hline
SST-5 & Neutral & Neutral\\
\hline
Amazon-2 & Positive & Positive\\
\hline
Amazon-5 & Positive & Positive\\
\hline
%\multicolumn{3}{l}{$^{\mathrm{b}}$OP stands for Overall Polarity}
\end{tabular}
\label{tab5}
\end{center}
%\textbf{Table \ref{tab5}} shows overall polarity computation for all the datasets. The Original OP is the same as the Computed OP for all the datasets.
\end{table}

\section{CONCLUSION}\label{concl}
In this section, we consolidate our efforts by offering a succinct summary of our work, highlighting the contributions we have made to the domain's knowledge, and outlining prospective avenues for future exploration based on the insights gleaned from our observations.

\subsection{Conclusion} \label{conclusion}

In this endeavor, we have expanded the existing domain knowledge of SA through our primary contributions, showcasing significant advancements:

First, we have adeptly fine-tuned BERT, a pivotal aspect enhancing accuracy within movie reviews SA. Leveraging transfer learning, we coupled BERT with BiLSTM, utilizing BERT-generated word embeddings as inputs for BiLSTM. This amalgamation was instrumental for polarity and fine-grained classification tasks, spanning three-class, four-class, and five-class categorizations across various datasets, i.e., IMDb, MR, SST, and Amazon. Notably, our model consistently outperformed prior works across all classification tasks and datasets.

Moreover, to augment the accuracy of our model for five-class classification, we delved into the impact of employing SMOTE and NLPAUG on SST-5, a notably challenging fine-grained classification benchmark. Intriguingly, SMOTE led to a decrease in accuracy from 58.44\% to 58.36\%, while NLPAUG remarkably boosted accuracy to 60.48\%.

Second, we introduced a heuristic algorithm tailored to the BERT-BiLSTM output vector, dynamically adapting to the specific classification task at hand. Demonstrating its reliability, we confirmed that the original overall polarity aligned perfectly with the computed overall polarity across all datasets. Furthermore, variations within dataset versions exhibited consistent computed overall polarity.

This work marks a pioneering effort, coupling BERT with BiLSTM and applying the resulting model across diverse sentiment classification tasks and benchmark datasets. Notably, it is the first to utilize the output vector of a model for computing overall sentiment polarity. Our exploration not only enhances understanding regarding review polarity but also sheds light on the nuanced shifts in review and overall polarities as classification granularity intensifies. These combined contributions significantly advance the understanding and application of SA within diverse contexts.

\subsection{Future Work} \label{future_work}
Moving forward, our proposed future endeavors revolve around addressing key challenges in the realm of SA, aiming to enhance model performance and extract deeper insights. One significant area of focus involves exploring strategies to effectively apply accuracy improvement techniques to transformed BERT features, despite the inherent loss of semantic information during their transformation. This endeavor seeks to overcome the limitations posed by the semantic information loss, potentially revolutionizing the effectiveness of these techniques.

Additionally, our exploration will delve into the nuanced contributions of different sentence components to sentiment prediction. This unexplored facet holds immense potential, as the intricate interplay between sentence elements often remains untapped by current methodologies. By dissecting these contributions, we aim to unravel hidden layers of information critical to sentiment prediction, thereby enriching the understanding of sentence structures and their impact on SA.

Furthermore, our future roadmap includes an extensive exploration of alternative pre-trained language models beyond BERT, such as RoBERTa and GPT. This exploration aims to broaden the horizons of our analysis, leveraging the unique capabilities and architectures of these models to potentially enhance SA outcomes. Diversifying our approach by incorporating these SOTA models could unlock new dimensions and avenues for deeper exploration within the field of SA.

\section*{ACKNOWLEDGMENT}
We would like to express our gratitude to anonymous reviewers for their valuable insights and comments. We also acknowledge the Data Analytics that are Robust and Trusted
(DART) project under the National Science Foundation (NSF) for financially supporting this work. We also extend our gratitude to all individuals who contributed to this work.

\IEEEtriggeratref{40}
\bibliographystyle{IEEEtran}
\bibliography{mybib}

%\newpage
\end{document}